\documentclass{article}
\usepackage{spconf,amsmath,graphicx}
\usepackage{url}
\usepackage{xcolor}
\usepackage{multirow}
\usepackage{booktabs}
\usepackage{xcolor}

\def\d{{\mathbf d}}
\def\r{{\mathbf r}}

\newcommand{\rev}[1]{\textcolor{black}{#1}}

\DeclareMathOperator*{\argmin}{argmin}

\title{Speech Privacy Leakage from Shared Gradients in Distributed Learning}
\name{Zhuohang Li$^1$, Jiaxin Zhang$^2$, Jian Liu$^1$
}
\address{$^1$University of Tennessee, Knoxville \qquad $^2$Intuit AI Research}
\begin{document}
\maketitle
\begin{abstract}
Distributed machine learning paradigms, such as federated learning, have been recently adopted in many privacy-critical applications for speech analysis.
However, such frameworks are vulnerable to privacy leakage attacks from shared gradients.
Despite extensive efforts in the image domain, \rev{the exploration of speech privacy leakage from gradients is quite limited.}
In this paper, we explore methods for recovering private speech/speaker information from the shared gradients in distributed learning settings. We conduct experiments \rev{on a keyword spotting model with two different types of speech features} to quantify the amount of leaked information by measuring the similarity between the original and recovered speech signals. We further demonstrate the feasibility of inferring various levels of side-channel information, including speech content and speaker identity, under the distributed learning framework without accessing the user's data.
\end{abstract}
\vspace{-2mm}
\section{Introduction}
\label{sec:intro}

Voice assistants, such as Google Assistant, Amazon Alexa, and Apple Siri, have been widely deployed on various smartphones and smart speakers, as they provide a natural and convenient way for user interaction. The modern voice user interface is powered by deep neural networks, which enables efficient speech processing for many tasks, such as \textit{automatic speaker verification} (ASV)~\cite{snyder2017deep} and \textit{automatic speech recognition} (ASR)~\cite{amodei2016deep}. The remarkable performance of such models is fueled by a growing amount of training data; yet collecting data from users is becoming increasingly difficult due to privacy regulations~\cite{GDPR,CCPA} and user privacy concerns.

Distributed machine learning, which allows multiple data holders to jointly train a machine learning model under the coordination of a central server, has attracted great research attention.
Compared with the conventional centralized learning framework, data parallelism distributed training not only scales better to larger data sizes, but also provides a level of privacy to participating users by encoding the data minimization principle: clients are allowed to keep their private data on the local device and only share the gradient information to the server for updating the model. As such, distributed learning, especially the emerging \textit{federated learning} (FL), has been quickly deployed in production for many speech-related tasks, such as speaker verification~\cite{granqvist2020improving} and keyword spotting~\cite{hard2022production}.

\begin{figure}[t]
    \centering
    \includegraphics[width=0.9\linewidth]{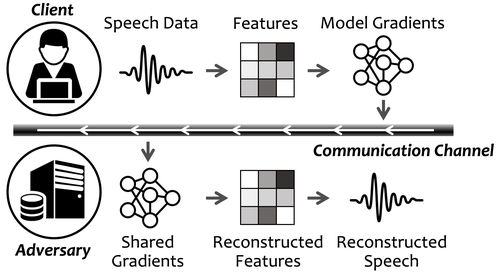}
    \vspace{-3mm}
    \caption{Illustration of speech privacy leakage from shared gradients in the distributed learning scenario.}
    \vspace{-3mm}
    \label{fig:overview}
\end{figure}

Recently, several studies~\cite{zhu2019deep,geiping2020inverting,yin2021see,li2022auditing} have revealed that image data can be recovered to some extent through the shared gradients in distributed learning (known as \textit{gradient leakage}, or \textit{gradient inversion}), which may pose severe threats to user's data privacy. However, to date, there has been \rev{limited} exploration of gradient leakage in the speech domain. Compared with image data, speech recordings are a rich source of personal and sensitive information that can be used to support a wide spectrum of applications, from sentiment analysis to spoken language recognition, and further to voice biometric profiling. Therefore, 
distributed learning involving speech data should be evaluated carefully to fully understand its potential privacy vulnerabilities.

To fill this gap, this work investigates the risk of gradient leakage on speech data with the following questions:
\begin{enumerate}
    \vspace{-1mm}
    \item How to recover private speech data from the shared gradients, if feasible?
    \vspace{-2mm}
    \item What level of private information (e.g., speech content, speaker identity, etc.) can be exposed from gradients?
    \vspace{-1mm}
\end{enumerate}

To answer the first question, we extend on previous gradient leakage attacks in the image domain and provide a two-stage inversion method that can numerically restore the speech waveform from the gradients shared by the client, which is illustrated in Fig.~\ref{fig:overview}.
We find that different from the image domain, deep learning models for processing speech data are usually trained on condensed speech features rather than the raw waveform. As a result, inverting the gradients only recovers the features of the original signal.
Moreover, unlike the image data which has well-defined value space for each pixel, the spectral/cepstral feature of speech signals has a wider range and is more prone to subtle errors which would be amplified when projecting back to the time domain, and thus would require more careful treatment.

To answer the second question, we design and conduct extensive experiments to investigate the amount of information that can be recovered through the gradients.
We explore a distributed learning scenario of speech command recognition on two types of features, i.e., Mel-spectrogram and Mel-frequency cepstral coefficients (MFCC), and utilize a suite of $4$ different metrics to quantify the recovered speech quality and the intelligibility of the recovered speech content.
Moreover, we further inspect the amount of leaked voice biometric information leveraging a pre-trained speaker verification model.
Our results show that compared with MFCC, using Mel-spectrogram as the front-end feature would lead to more leakage in speech content and speaker information of the client's private speech data from the shared gradients.
These findings can help the community understand the potential of gradient leakage on different speech features under the gradient sharing framework and design distributed learning schemes with enhanced privacy protection.

\vspace{-2mm}
\section{Related Work}
\label{sec:related}

\textbf{Distributed Learning for Speech}. Recently, an increasing amount of research effort has been put into utilizing federated learning as a privacy-enhancing technique to improve neural networks on distributed user devices, such as smartphones and smart speakers. Besides early studies~\cite{leroy2019federated,hard2020training,guliani2021training} conducted in simulated environments, many federated learning frameworks for speech processing have already been deployed in production, including Apple's speaker verification model~\cite{granqvist2020improving} and Google's keyword spotting model~\cite{hard2022production}. As human speech contains rich semantics of sensitive and personal information, there is a pressing need for thoroughly understanding the privacy risks of models trained on speech data in a distributed manner. 

\noindent \textbf{Privacy Leakage from Gradients}. A few recent studies have shown that sharing gradients in distributed learning is not as safe as it has been presumed. Zhu \textit{et al.}~\cite{zhu2019deep} first demonstrated that it is possible to recover clients' private images from the shared gradients by generating fake images that minimize the gradient matching loss. Geiping \textit{et al.}~\cite{geiping2020inverting} extended this method to deeper neural networks with higher resolution images with a different loss function design. Following work~\cite{yin2021see,li2022auditing} further improved on this by exploring various prior information.
Compared to the remarkable progress in the image domain, there is a lack of research on the feasibility and severity of gradient leakage in speech data. \rev{To our knowledge, the most relevant study by Dang \textit{et al}.~\cite{dang2022method} proposed a method to infer speaker identity from gradients of an ASR model. However, such a method only recovers the speech features instead of the original speech waveform. Moreover, compared with ASR models that are trained on long spoken sentences, the lightweight keyword spotting models for recognizing speech commands are more commonly deployed in distributed learning setting, yet the privacy risk there is still unexplored.} This work is devoted to filling this research gap.

\vspace{-2mm}
\section{Methodology}
\label{sec:method}
\vspace{-2mm}

\subsection{Problem Formulation}

We consider a supervised learning task under the canonical distributed learning setting that involves two parties: the server $S$ and the clients $C$. The learning objective is to optimize the parameters $\theta$ of a neural network $f_\theta$ to minimize the empirical risk measured by loss function $\mathcal{L}$ on all training data:
$
    \min_{\theta} \sum_{c \in C} \sum_{(x_i, y_i) \in \mathcal{D}_c} \mathcal{L}(f_\theta(x_i), y_i),
$
where $x_i$, $y_i$ are the local data and label from client $c$'s local dataset $\mathcal{D}_c$. Instead of directly sharing their private data, federated learning allows participating clients to only shared the gradients computed on their private data, i.e., $\nabla_{\theta} \mathcal{L}(f_{\theta}(x_i), y_i)$. The server then collects and aggregates gradients from all participating clients to update the global model for each communication round.

\noindent \textbf{Threat Model}. We assume the adversary cannot interfere with the normal federated learning procedure but has access to the gradients uploaded by each individual client. In practice, the adversary can be an honest-but-curious server or a malicious analyst that eavesdrops on the communication channel.

\noindent \textbf{Objective}. The adversary's objective is to infer private information about the client by attempting to reconstruct the client's private data. Previous studies~\cite{zhu2019deep,geiping2020inverting} have shown that this can be done through generating synthetic data samples to match the client's gradients. Let $\Delta \theta$ denote the actual gradients shared by the client and $x', y'$ represent the synthetic data sample and label. Formally the adversary solves for:
\begin{equation}\label{eq:objective}
\setlength{\belowdisplayskip}{1pt}
    x'^*, y'^* = \argmin_{x', y'} \d(\Delta \theta, \nabla_{\theta} \mathcal{L}(f_\theta(x'), y')),
\end{equation}
where $x'^*, y'^*$ is the minimizer of the distance between gradients measured by a distance metric $\d$. Intuitively, if the gradients computed on the synthetic data closely match the actually shared gradients, the synthetic data will also recover the important semantics of the client's private data. In practice, as the label $y'^*$ can be analytically restored from the gradients~\cite{geiping2020inverting,yin2021see}, the adversary only needs to solve for $x'^*$.

\vspace{-2mm}
\subsection{Recovering Speech Data From Gradients}
\textbf{Challenge}. In the image domain, deep learning models are usually trained to directly process raw imagery data. Differently, in the speech domain, it is a common practice to first extract spectral or temporal acoustic features (e.g., spectrogram or cepstral coefficients) from the raw speech signal and then feed the extracted features into the deep learning model. This creates an additional layer of difficulty to gradient inversion since solving Eq.~\ref{eq:objective} only recovers the features rather than the original speech waveform.

\noindent \textbf{Method}. To address the above challenge, the proposed method recovers speech data from the shared gradients in the following two stages:

(1) \textit{Feature Reconstruction}. The goal of the first stage is to recover the acoustic features from the gradients shared by the users. Specifically, let $u$ denote the set of 2-dimensional spectral features extracted from speech waveform $x$. We then solve the optimization problem in Eq.~\ref{eq:objective} by minimizing the Euclidean distance in the model parameter (gradient) space:
\begin{equation}\label{eq:loss}
\setlength{\belowdisplayskip}{1pt}
    u'^* = \argmin_{u'} ||\Delta \theta - \nabla_{\theta} \mathcal{L}(f_\theta(u'), y'^*)||^2_2 + \lambda \r(u'),
\end{equation}
where $\r$ is a regularization term and $\lambda > 0$ is a weighting parameter. In this work, we use anisotropic total variation~\cite{lou2015weighted} as the regularizer, i.e., $\r(u') = ||d_h u'||_1 + ||d_v u'||_1$, where $d_h$, $d_v$ denote the horizontal and vertical partial derivative operators, respectively. We do not bound the search space for $h'$ during optimization since unlike image data, the values of the acoustic features do not reside in a well-defined range.

(2) \textit{Waveform Reconstruction}. The goal of the second stage is to reconstruct the speech waveform based on the features recovered from the first stage. In this work, we consider recovering the waveform from two common types of features for speech processing: Mel-spectrogram and MFCC. To convert a Mel-spectrogram back to a time-domain signal, we first create Mel filter banks and then approximate the normal short-time Fourier transform (STFT) magnitudes by searching for the non-negative least squares solution that minimizes the Euclidian distance between the target Mel-spectrogram and the product of the estimated spectrogram and the filter banks. Then the Griffin-Lim algorithm~\cite{griffin1984signal} is applied to produce the speech waveform by estimating the missing phase information. As for MFCCs, an extra step is needed to first invert the cepstral coefficients to approximate a Mel-spectrogram. This is done by first applying the inverse discrete cosine transform (iDCT) and then map the decibel-scaled results to a power spectrogram\footnote{\url{https://librosa.org/doc/main/generated/librosa.feature.inverse.mfcc_to_mel.html}}. After that, the regular Mel-spectrogram inversion procedure is applied to further get the estimated waveform.

\vspace{-2mm}
\section{Experiments}
\label{sec:exp}

\subsection{Experimental Setting}

\begin{table}
\centering
\caption{Model used in evaluation.}\label{tab:model}
\vspace{1.5mm}
\resizebox{0.7\linewidth}{!}{
\begin{tabular}{cccc}
    \hline \hline
    Type     & Kernel & Stride & Output \\ \hline
    Conv2D       & (3, 3)       & (1, 1)       & (30, 30, 32)    \\
    Conv2D       & (3, 3)       & (1, 1)       & (28, 28, 64)    \\
    MaxPooling2D       & (2, 2)       & (2, 2)       & (14, 14, 64)    \\
    Flatten     &  --      &   --     & 12544     \\
    FC &  --   &  --  & 128       \\ 
    FC &  --   &  --  & 10       \\ \hline \hline
    \end{tabular}
}
\vspace{-2mm}
\end{table}

\noindent \textbf{Dataset}.
We use speech data from the Speech Commands dataset~\cite{warden2018speech} to conduct evaluation. The dataset was developed for developing and testing compact and effcient on-device keyword spotting model, which is one of the most widely-adopted federated learning applications in production, and thus is well-suited for our task. Each sample of the dataset contains $1$ second recording of spoken speech commands sampled at $16$kHz and corresponding label. We select a subset of the dataset that contains $10$ common words (i.e., ``yes'', ``no'', ``up'', ``down'', ``left'', ``right'', ``on'', ``off'', ``stop'', and ``go'') as in the first released version of the dataset.

\noindent \textbf{Front-end Feature Extraction}.
To extract acoustic features, the speech signal is first pre-emphasised with a factor of $0.97$. Then speech frames are created using overlapping Hamming
windows with length of $2,048$ samples and frame-shift of $512$ samples. For \textit{Mel-spectrogram}, 512 point fast Fourier transform (FFT) with $32$ Mel bands is used. For \textit{MFCC}, $128$-channel filterbank is used to extract $32$-dimensional coefficient features which are further processed by cepstral mean and variance normalization (CMVN).

\noindent \textbf{Distributed Learning Setting}.
(1) \textit{Model}.
We adopt the same model structure as used in the Tensorflow keyword recognition example\footnote{\url{https://www.tensorflow.org/tutorials/audio/simple_audio}}.
The model is composed of two 2D convolution layers, one max-pooling layer, and two fully-connected layers. The detailed model architecture is described in Table~\ref{tab:model}.
(2) \textit{Gradient Computation}.
We assume the clients compute one local step of gradient descent on one random sample from their private dataset and then send back the gradients (i.e., distributed stochastic gradient descent).

\noindent \textbf{Parameters}. We solve Eq.~\ref{eq:loss} using Adam optimizer for $8,000$ iterations with $\lambda=0.001$ and a learning rate of $0.01$. To avoid local minimum, each sample is given $2$ trials and the reconstruction result with the lower loss is selected.

\vspace{-2mm}
\subsection{Experimental Results}

\subsubsection{Reconstruction Evaluation}
To evaluate the reconstruction performance, we compare the resulting speech signal recovered from gradients using our method with the original signal using the following metrics:

\begin{table}
\centering
\caption{Reconstruction results on $400$ speech samples.} \label{tab:stat}
\vspace{1.5mm}
\resizebox{\linewidth}{!}{
\begin{tabular}{c|c|c|c|c} 
\hline\hline
& \textbf{F-MSE $\downarrow$} & \textbf{W-MSE $\downarrow$} & \textbf{PESQ $\uparrow$} & \textbf{STOI $\uparrow$} \\
\hline\hline
\multicolumn{5}{c}{\textit{Inverting from \textbf{Features}}} \\ \hline
\textbf{Mel-spectrogram}           & \begin{tabular}[c]{@{}c@{}}$-$\end{tabular}                         & \begin{tabular}[c]{@{}c@{}}0.0097\\$\pm$ 0.0130\end{tabular}                      & \begin{tabular}[c]{@{}c@{}}2.1090\\$\pm$ 0.3563\end{tabular}               & \begin{tabular}[c]{@{}c@{}}0.8082\\$\pm$ 0.0715\end{tabular}                          \\ 
\hline
\textbf{MFCC}           & \begin{tabular}[c]{@{}c@{}}$-$\end{tabular}                         & \begin{tabular}[c]{@{}c@{}}0.0091\\$\pm$ 0.0119\end{tabular}                      & \begin{tabular}[c]{@{}c@{}}2.1041\\$\pm$ 0.3762\end{tabular}               & \begin{tabular}[c]{@{}c@{}}0.7856\\$\pm$ 0.0760\end{tabular}                          \\ 
\hline
\multicolumn{5}{c}{\textit{Inverting from \textbf{Gradients}}} \\ \hline
\textbf{Mel-spectrogram}           & \begin{tabular}[c]{@{}c@{}}0.0002\\$\pm$ 0.0017\end{tabular}                         & \begin{tabular}[c]{@{}c@{}}0.0095\\$\pm$ 0.0129\end{tabular}                      & \begin{tabular}[c]{@{}c@{}}2.0427\\$\pm$ 0.3816\end{tabular}               & \begin{tabular}[c]{@{}c@{}}0.8004\\$\pm$ 0.0804\end{tabular}                          \\ 
\hline
\textbf{MFCC}           & \begin{tabular}[c]{@{}c@{}}5701.6225\\$\pm$ 1919.7741\end{tabular}                         & \begin{tabular}[c]{@{}c@{}}0.0153\\$\pm$ 0.0086\end{tabular}                      & \begin{tabular}[c]{@{}c@{}}1.3886\\$\pm$ 0.1750\end{tabular}               & \begin{tabular}[c]{@{}c@{}}0.4259\\$\pm$ 0.1234\end{tabular}                          \\ 
\hline\hline
\end{tabular}
}
\vspace{-2mm}

\end{table}

\noindent \textbf{Evaluation Metrics}.
(1) \textit{F-MSE}: the mean squared error between the ground truth features of the original speech and the reconstructed features.
(2) \textit{W-MSE}: the mean squared error between the original speech waveform and the reconstructed waveform.
(3) \textit{PESQ}: the perceptual evaluation of speech quality (PESQ)~\cite{beerends2002perceptual} score is designed for end-to-end quality assessment of degraded audio sample in narrow-band telephone networks. The computed score is in the range of $[-0.5, 4.5]$, with higher scores indicating better speech quality.
(4) \textit{STOI}: the short-time objective intelligibility (STOI)~\cite{taal2011algorithm} metric measures the intelligibility of degraded speech signals based on a correlation coefficient between the temporal envelopes of the reference and degraded signals in short-time overlapping segments.

\begin{figure}[t]

\begin{minipage}[b]{0.9\linewidth}
  \centering
  \centerline{\includegraphics[width=\linewidth]{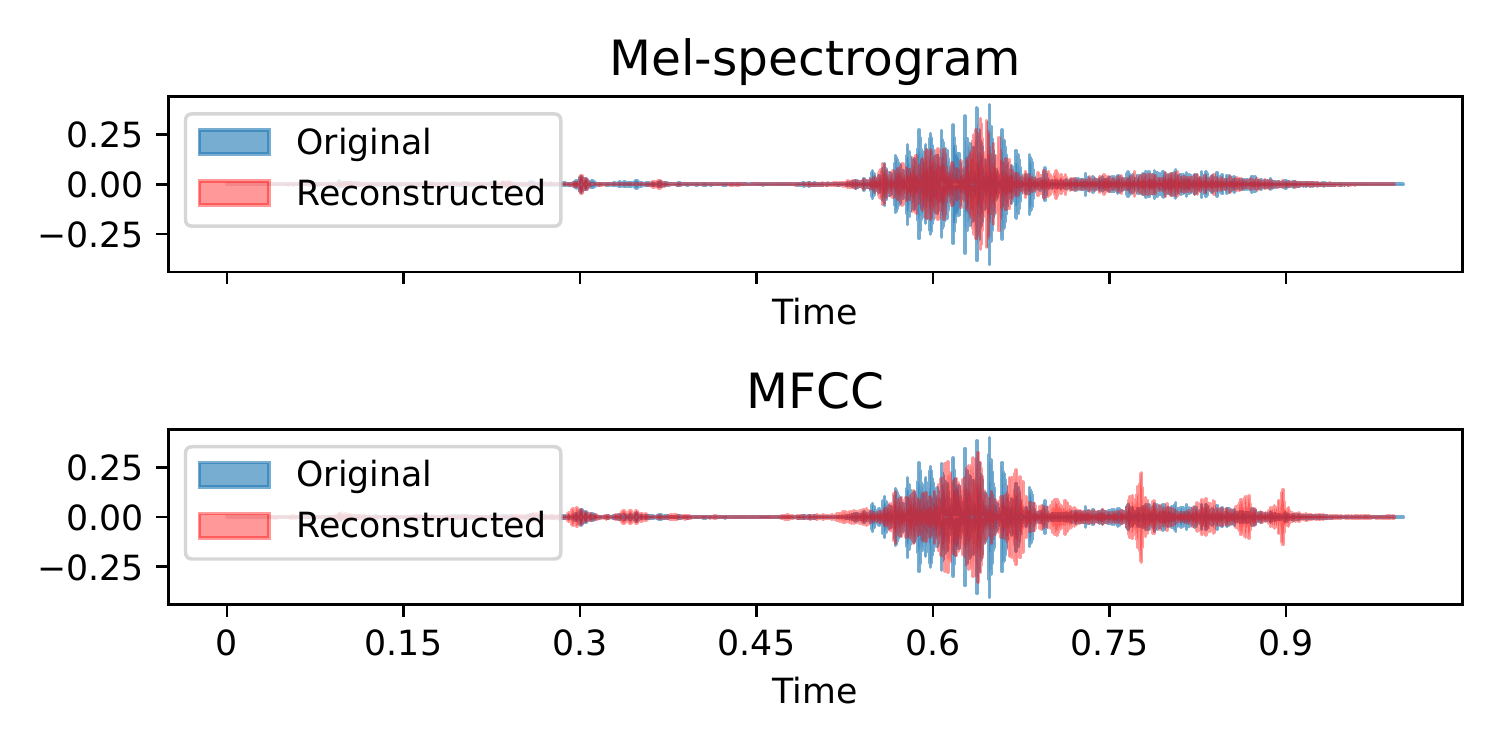}}
 \vspace{-2mm}
  \centerline{(a) Waveform}\medskip
 \vspace{-1mm}
\end{minipage}
\begin{minipage}[b]{0.935\linewidth}
  \centering
  \centerline{\includegraphics[width=\linewidth]{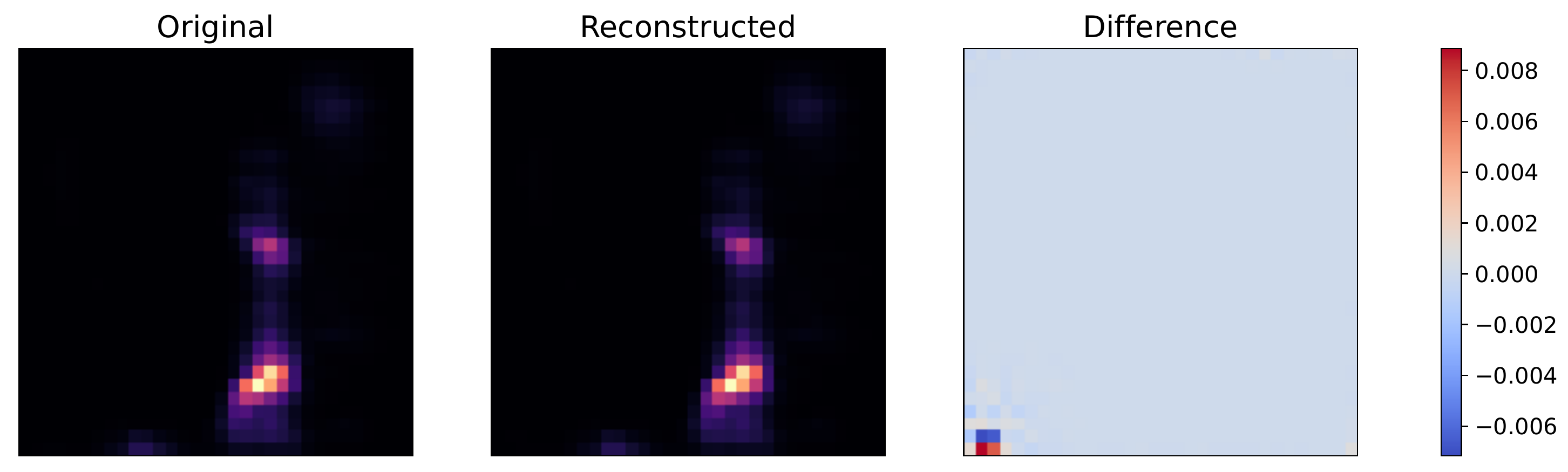}}
 \vspace{-1mm}
  \centerline{(b) Mel-spectrogram}\medskip
 \vspace{-2mm}
\end{minipage}

\begin{minipage}[b]{0.92\linewidth}
  \centering
  \centerline{\includegraphics[width=\linewidth]{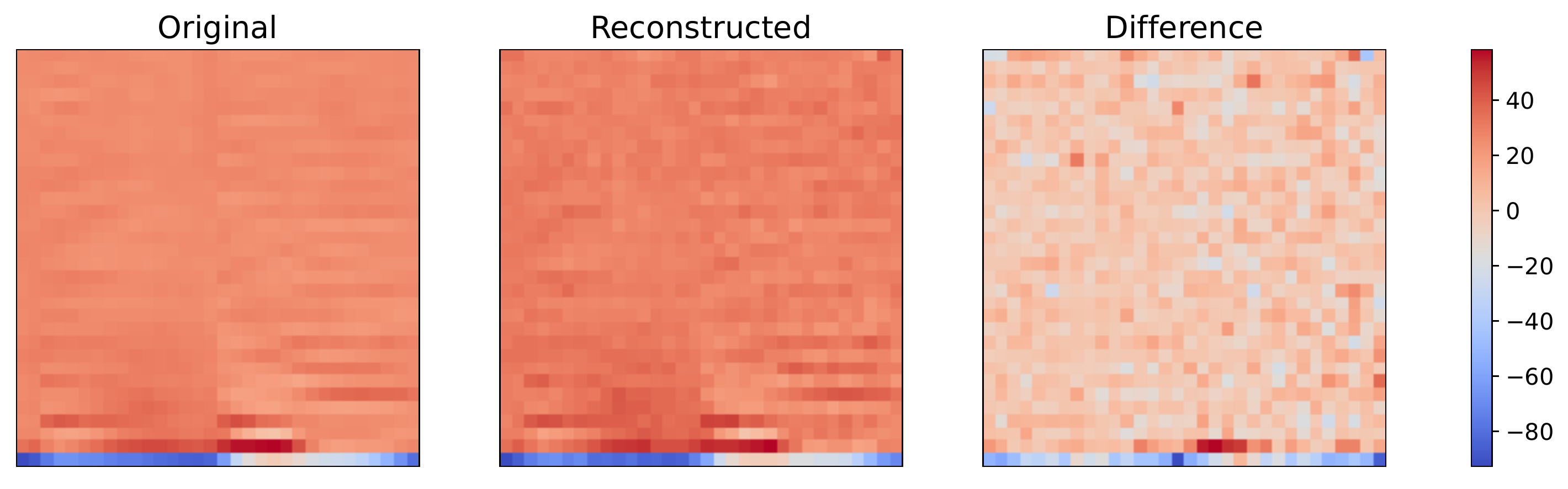}}
 \vspace{-1mm}
  \centerline{(c) MFCC}\medskip
 \vspace{-1mm}
\end{minipage}
 \vspace{-2mm}
\caption{Visualizing reconstructed speech command ``yes''.}
 \vspace{-2mm}
\label{fig:vis}
\end{figure}

\noindent \textbf{Results}.
Table~\ref{tab:stat} presents the statistical results of conducting the reconstruction on $400$ random speech samples from the testing set. We also show the results of inverting from the ground truth features (i.e., only conducting waveform reconstruction) as baseline for comparison.
An example of visualizing the reconstructed speech command ``yes'' is provided in Fig.~\ref{fig:vis}.
We observe that for Mel-spectrogram, inverting from gradients yields a similar performance as directly inverting from features, and the reconstructed waveform is very close to the original waveform, with measured W-MSE $<0.001$ and PESQ $>2$. However, for MFCC, inverting from gradients would induce a large distortion, causing the quality of the reconstructed speech to degrade drastically. This is potentially because the coefficient values are in decibel scale and has high variance and thus are prone to small perturbations, which makes it harder to launch gradient leakage attacks against MFCCs.

\vspace{-2mm}
\subsubsection{Speaker Re-identification}

\begin{table}
\centering
\caption{Speaker re-identification results on $400$ speech samples.}\label{tab:spk}
\vspace{1.5mm}
\resizebox{0.82\linewidth}{!}{
\begin{tabular}{c|c|c} 
\hline \hline
                & \textbf{Score $\uparrow$}                              & \textbf{Success Rate $\uparrow$}  \\ 
\hline \hline
\multicolumn{3}{c}{\textit{w.r.t. \textbf{Inverted Signal from Features}}}           \\ 
\hline
\textbf{Mel-spectrogram} & 0.7288 $\pm$ 0.1386 & 99.25\%          \\ 
\hline
\textbf{MFCC}            & 0.1574 $\pm$ 0.1119 & 16\%           \\ 
\hline
\multicolumn{3}{c}{\textit{w.r.t. \textbf{Original Signal}}}                         \\ 
\hline
\textbf{Mel-spectrogram} & 0.4445 $\pm$ 0.1450 & 90.5\%           \\ 
\hline
\textbf{MFCC}            & 0.0514 $\pm$ 0.0916 & 2.5\%           \\
\hline \hline
\end{tabular}
}
\vspace{-2mm}

\end{table}

To examine whether the speaker information (i.e., voice biometric) can be retained through the reconstruction, we pass the signal recovered from gradients and the reference signal into a speaker verification model and report the cosine similarity score of the embeddings and the success rate of the two signals being recognized as the same speaker. To perform speaker verification, we use the ECAPA-TDNN model~\cite{desplanquesTD20} pretrained on Voxceleb dataset provided by SpeechBrian\footnote{\url{https://huggingface.co/speechbrain/spkrec-ecapa-voxceleb}}.

\noindent \textbf{Results}.
Table~\ref{tab:spk} presents the statistical results of conducting speaker re-identification on the same set of $400$ speech samples. For comparison, we show the results measured w.r.t. both the inverted signal from the ground truth features and w.r.t. the original signal. We observe that the speech signals reconstructed from Mel-spectrogram preserve most speaker information, with $99\%$ and $90\%$ chance to pass the speaker verification when compared to the inverted signal and original signal, respectively. In contrast, speech signals reconstructed from MFCC have a very low probability to be verified as the same speaker, especially when directly compared to the original signal.

\vspace{-2mm}
\section{Conclusion}
\label{sec:conclusion}
\vspace{-1mm}

In this work, we study the potential of the privacy leakage of speech data from shared gradients by proposing a two-stage inversion method that sequentially achieves speech feature reconstruction from gradients and waveform reconstruction from the recovered features.
Through extensive experiments, we demonstrate that compared to Mel-spectrogram, MFCC exhibits better resilience against gradient leakage attacks, with less leaked speech/speaker information.
Future work could investigate neural vocoders for better waveform reconstruction quality.

\vfill
\pagebreak

\bibliographystyle{IEEEbib}
\bibliography{refs}

\end{document}